# Effective and Acceptable Eco-Driving Guidance for Human-Driving Vehicles: A Review

**Ran Tu (Corresponding Author)**
turancoolgal@seu.edu.cn
School of Transportation, Southeast University, Nanjing, China

**Junshi Xu**
junshi.xu@mail.utoronto.ca
Department of Civil and Mineral Engineering, University of Toronto, Toronto, Canada

# Effective and Acceptable Eco-Driving Guidance for Human-Driving Vehicles: A Review

## ABSTRACT

Eco-driving guidance includes courses or suggestions for human drivers to improve driving behaviour, reducing energy use and emissions. This paper presents a systematic review of existing eco-driving guidance studies and identifies challenges to tackle in the future. A standard agreement on the guidance design has not been reached, leading to difficulties in designing and implementing eco-driving guidance for human drivers. Both static and dynamic guidance systems have a great variety of guidance results. In addition, the influencing factors (such as the suggestion content, the displaying methods, and drivers' socio-demographic characteristics) have opposite effects on the guidance result across studies, while the reason has not been revealed. Drivers' motivation to practice "eco" behaviour, especially long-term, is overlooked. Besides, the relationship between users' acceptance and system effectiveness is still unclear. Adaptive driving suggestions based on drivers' habits can improve the effectiveness, while this field is under investigation.

## 1. INTRODUCTION

On-road transportation is an essential source of greenhouse gas (GHG) and air pollutant emissions among all economic sectors (Henneman et al., 2021; Ministry of Ecology and Environment of the People's Republic of China, 2021; The Government of Canada, 2021). Numerous studies have investigated the factors that influence the amount of vehicle emissions and fuel consumption, and proposed control strategies accordingly. Compared to other methods such as automated electric vehicles and traffic demand control, adopting ecological driving behaviour (eco-driving) is a more cost-effective method because it does not require the upgrade of the vehicle powertrain or the reconstruction of infrastructure. Instead, vehicle operational emissions and fuel consumption can be decreased by correcting driving trajectory.

A large number of eco-driving studies focus on the trajectory control of vehicles through Connected and Autonomous Vehicle (CAV) technologies. Scenario-specific studies were proposed to optimize speed and acceleration profiles when vehicles were approaching signalized intersections or driving through highway links. Look-ahead traffic condition, signal control information, and road geometric designs are delivered to CAVs through the V2X (Vehicle to X communication), and vehicle speed can be dynamically adjusted with the objective of the optimal fuel consumption (Yang H. et al., 2017; Yang X.F. et al., 2020; Zhai et al., 2019). However, the speed-time profile determined by eco-driving algorithms cannot be practiced precisely when vehicles are operated by human drivers. Therefore, how to effectively guide drivers to modify their aggressive driving behaviour thus approaching the optimal driving operation is an important step to eco-driving. The definition of eco-driving behaviour is varied among previous studies. In general, it can be summarized as the following "Golden Rules" (ecodrive.org, 2013; Fu et al., 2019; Orfila et al., 2015):

(a) Keep a steady speed

(b) Anticipating surrounding traffic flow

(c) Braking smoothly

(d) Reducing long-time idling

(e) Avoiding overuse of auxiliary devices

The "Golden Rules" provide basic guidelines for eco-driving. In the real world, however, due to diverse drivers' habits and varied traffic conditions, more practical eco-driving guidance should be provided on how to operate the vehicle properly with the purpose of energy-saving and emission reduction, instead of strategic guidelines.

The objective of this paper is to present an overview of existing studies on eco-driving guidance for human-driving vehicles, with focus on the effectiveness and acceptance of the guidance. Methods to identify eco-driving behaviour, different presence formats of the guidance, their effects on drivers' behaviour, and associated influencing factors are discussed. Based on the overview, this paper concludes challenges and research gaps of existing studies, highlighting efforts that can be made in the future to develop an effective and acceptable eco-driving guidance system.

## 2. METHODOLOGY

A systematic review process is conducted based on the PRISMA guidance (Cooke et al., 2012). The SPIDER search analysis (Cooke et al., 2012) is used to determine the eligibility criteria for the review.

**Sample (S):** All categories of drivers, including novice and experienced drivers; all types of human-driving vehicles, including fossil fuel-powered vehicles, electric vehicles, hybrid vehicles, as well as vehicles in various sizes

**Phenomenon of Interest (P.I.):** Eco-driving guidance design that aims to improve drivers' eco-driving skills, or improve drivers' acceptance.

**Design (D):** driving simulators or on-road driving experiments with eco-driving training or guidance provided

**Evaluation (E):** Emission reduction and energy saving. Drivers' acceptance level and the attitude towards the eco-driving guidance. Driver behavioural change.

**Research Type (R):** mixed methods.

The literature search was completed in the spring of 2021. Due to the broad inclusion of publishers, Google Scholar is selected as the database. In the starting phase of the search, keywords including "eco-driving" AND "training", "assistance", "feedback", "guidance", or "advice" were used in the search and the publication year was constrained between 2011 to 2021. By setting such thresholds, papers related to automated driving can be mostly excluded. 1040 papers met the criteria in the first round and entered the second-round selection. In the second round, papers that were written in English, published in peer-reviewed journals or peer-reviewed conferences, and with full-text access, were selected. We further implemented title and web-snapshot screening to exclude repetitive papers or papers without keywords in the manuscript. Noted we integrated the duplication removal with the title screening for two reasons: first, it is more convenient for the entire searching work; second, we noticed several papers have similar methodologies and results, while this type of repetition is hard to identify unless the title and web-snapshot screening are implemented. In this stage, 221 papers entered the next round. In the third-round filtering, an intense abstract screening is conducted to determine if the paper is eligible through the criteria of SPIDER analysis.

Finally, 77 papers investigating the effect and the drivers' acceptance of different types of eco-driving guidance design, are reviewed. The searching and filtering process are illustrated in Figure 1.

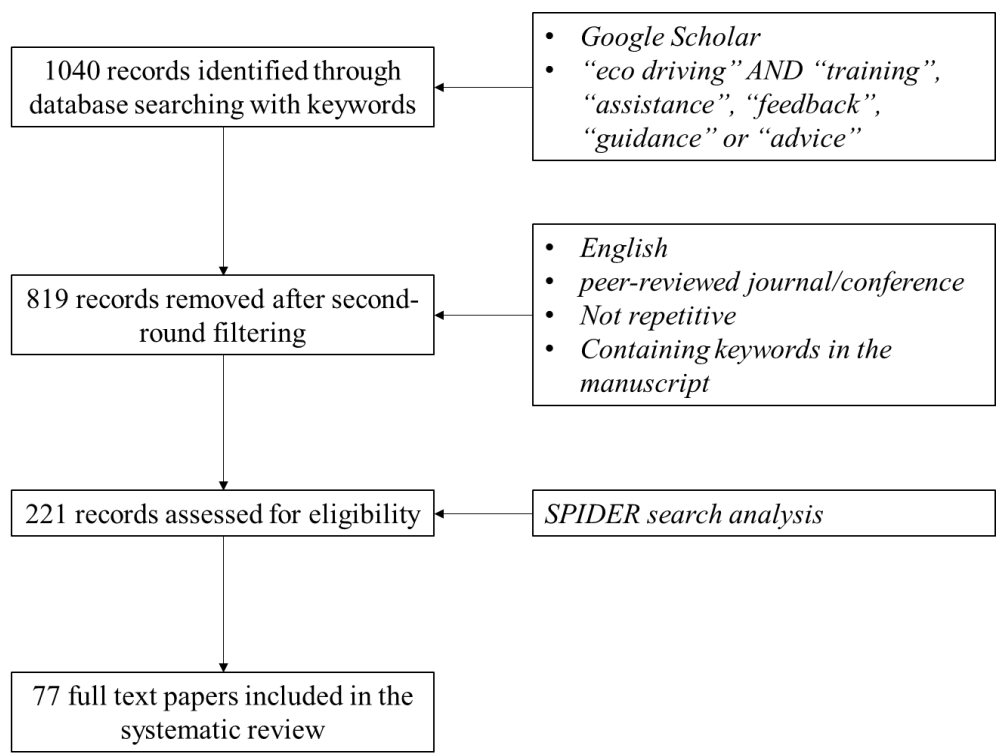

Figure 1 Article selection process in this systematic review

## 3. IDENTIFYING ECO-DRIVING BEHAVIOUR

Driving behaviour, such as gear shifting (for manual transmission), braking, accelerating, turning, and using auxiliary, influences the power demand of vehicles, and consequently affects fuel usage and emissions. Appropriate driving behaviour effectively leads to energy saving and emission reduction. Identifying eco-driving behaviour provides initial objectives that drivers are guided to accomplish through the eco-driving guidance.

Statistical and modal-based methods have been applied in previous research to explore driving operations that improve fuel efficiency. Díaz-Ramirez et al. (2017) applied a multiple linear regression model to identify the most influencing factors on fuel consumption. The result shows that engine speed is positively related to fuel usage. The multiple regression method was used in Lois et al. (2019), and the study found that compared to average speed, maximum speed is more important to identify eco-driving. In addition, this study confirms that eco-driving behaviour varies under different traffic conditions and road geometric designs. Based on the relationship between fuel consumption and driving behaviour under different traffic conditions, Wu et al. (2017a) summarized nine eco-driving related events that need to be appropriately controlled, including rapid acceleration, sharp deceleration, continuous acceleration for a long time, idling for a long time, lower driving speed, stop-and-go, appropriate cruising, proper start and stop operations. By distinguishing urban, suburban, and highway drive cycles, Ma et al. (2015) analyzed the relationship between fuel consumption and 26 driving characteristics related to speed, acceleration, deceleration, and gear shift. The study indicates that drivers' behaviour tends to be more influential in the normal running process where traffic condition is less congested.

In addition to deriving distinct eco-driving behaviour from the time-series driving record, recent studies have tried driving behaviour profiling to distinguish driving styles. Profiling means categorizing or rating drivers' behaviour based on specific criteria. The result can be presented as a categorical level, or a continuous numeric score, and a higher rating usually results from less energy consumption or fewer pollutant emissions (Singh and Kathuria, 2021a). The ranking of energy saving among all participating drivers is one of the most common approaches to calculate the eco-score (Chantranuwathana et al., 2014; Evin et al., 2020; Xu et al., 2021, 2020), whereas other studies have tried to define ecological levels (Yu et al., 2021), or to calculate a score through linear conversion using the amount of energy saving or emission reduction (Chen et al., 2018).

With recent studies revealing the positive relationship between aggressiveness and fuel consumption (Massoud et al., 2019; Xing et al., 2020), driving aggressiveness, or in other words, the data of driving behaviour, can be used alone to calculate eco-scores and represent the level of eco-driving. Driving aggressiveness is defined by various driving parameters, such as throttle position, braking pedal, speed, acceleration, jerk, and anticipation of traffic flow (Fitzpatrick et al., 2017; Singh and Kathuria, 2021b). This is more convenient than accessing the fuel consumption or emissions from external devices such as the OBD or portable emission measurement systems. Methods for the aggressiveness rating include rule-based classification (Javanmardi et al., 2017; Ouali et al., 2016; Prakash and Bodisco, 2019) and algorithm-based classification, such as fuzzy inference model (Castignani et al., 2013; Derbel and Landry, 2015; Massoud et al., 2018), unsupervised learning (Mantouka et al., 2019), probabilistic approaches (Huang et al., 2018), and Neural Network (Saleh et al., 2018). The rating can be used in the eco-driving guidance to stimulate the intrinsic motivation of drivers to improve their driving behaviour. In addition, specific driving suggestions can be provided by categorizing drivers' aggressiveness. The behavioural change, including the mean and standard deviation of acceleration/deceleration, average speed, and the position of the gas pedal or the brake pedal, as well as the anticipation to traffic, can be used to evaluate the effectiveness of proposed eco-driving guidance (for example, Hibberd et al.(2015),

A. H. Jamson et al. (2015), Wu et al. (2018, 2017b)).

# 4. TWO TYPES OF ECO-DRIVING GUIDANCE

Existing research presents two types of eco-driving guidance: static training and dynamic guidance. The difference between them is the use of real-time driving evaluation. The design of the system and the experiments conducted in existing studies are presented in Figure 2. For static training, driving suggestions are provided to drivers through training courses, videos, or education brochures. These suggestions are proposed based on the widely recognized definition of eco-driving, such as the "Golden Rules". Drivers are trained by self-learning or coaches, and then they apply the eco-driving knowledge to their daily driving activities. For dynamic guidance, however, driving suggestions are generated based on real-time driving behaviour analysis. Monitoring devices, such as Global Positioning System (GPS) and On-Board Diagnostics (OBD), are equipped in-vehicle. Instantaneous speed and acceleration trajectories are collected. Some experiments also record driving operations such as braking pedal, accelerator pedal, and gear shifting. In this case, energy consumption is usually recorded due to the accessibility to the CAN-BUS data. Analysis of energy consumption or emissions during the recorded period is then conducted, and corresponding driving suggestions (or warnings) can be generated. The name "dynamic" means that driving suggestions are subject to change with periodic evaluation of emission reduction or energy saving. In this section, two types of eco-driving guidance strategies and representative examples are discussed from aspects of their effectiveness and drivers' acceptance.

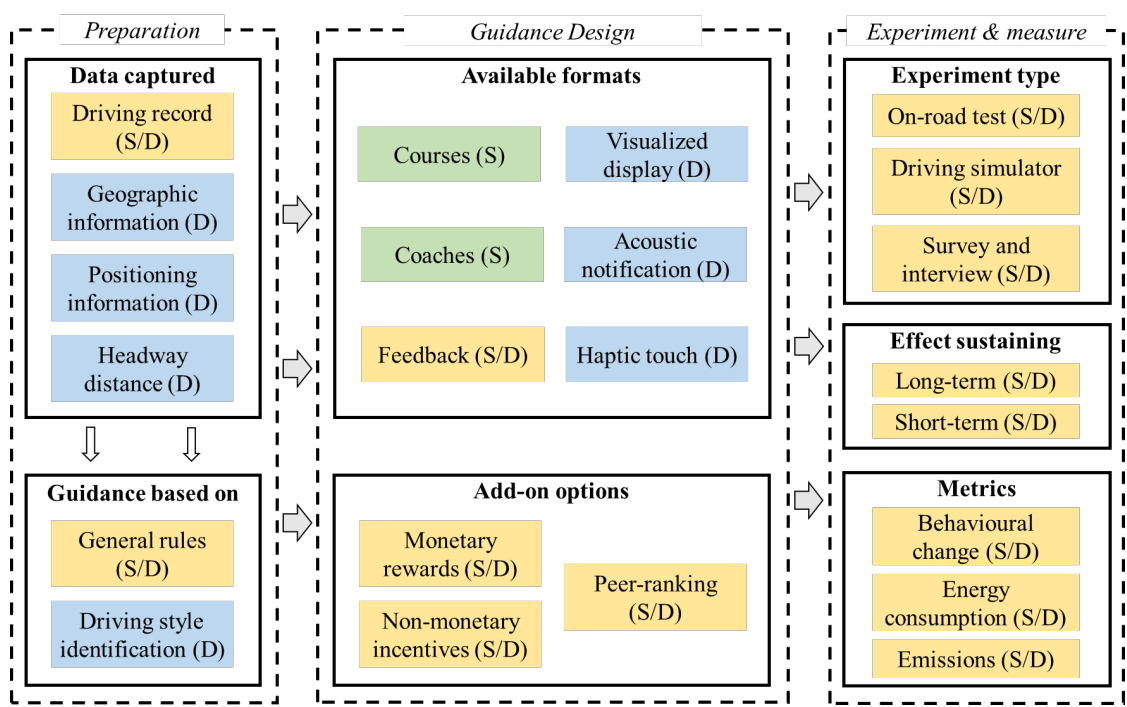

Figure 2 System elements and experiment design of two types of eco-driving guidance (where "S" means the item is only available in static guidance, "D" means the item is only available in dynamic guidance, "S/D" represents the item is applicable for both static and dynamic guidance)

## 4.1 Static eco-driving training based on pre-determined guidelines

Static eco-driving training provides driving suggestions based on wide-recognized eco-driving guidelines (such as the "Golden Rules") to drivers through training courses, video, brochures, and in-field practice with coaches. The effect of the training is evaluated by monitoring drivers' behaviour and comparing fuel consumption or emissions before/after the training session. Numerous eco-driving training programs have experimented in different jurisdictions worldwide (ecodrive.org, 2013; Ho et al., 2015; Jeffreys et al., 2018), and the change of driving behaviour is measured by comparing fuel consumption and emissions, or simply comparing key indicators of eco-driving, such as average speed, acceleration or deceleration rate, frequency of idling, and engine speed (rotation per minute, or RPM). Eco-driving training programs are affected by a wide range of factors, such as participating vehicle types, road geometry, drivers' socio-demographic characteristics, and the tested time span; therefore, their effects on behavioural change, energy-saving, and emission reduction greatly vary among programs. Table 1 summarizes reviewed articles targeting the energy/emission effect of static eco-driving guidance.

Table 1 Effects of static eco-driving guidance on energy saving and emission reduction* (ascending order of the published year)

| Study | Vehicle type** | Guidance type | Guidance design | Add-on options | Experiment type | Effect sustaining | Energy saving/emission reduction effects |
|-------|---------------|---------------|-----------------|----------------|-----------------|-------------------|------------------------------------------|
| (Rutty et al., 2013) | Medium class vehicles | Static | Courses | / | On-road test | Immediate | On average $CO_2$ reduction by 1.7kg per vehicle per day |
| (Andrieu and Pierre, 2014) | Light-duty vehicles | Static | Courses | / | On-road test | Immediate | On average 12% fuel saving |
| (Abuzo and Muromachi, 2014) | Private vehicles | Static | Courses | / | On-road test, survey | 1 week after the training | On average 0.894L to 1.378L of fuel saving per kilometer |
| (Rutty et al., 2014) | Resort vehicles | Static | Courses | / | On-road test | Five months | 8% fuel reduction, 8% $CO_2$ reduction |
| (Ho et al., 2015) | Private vehicles | Static | Courses, coaches | / | On-road test | Immediate | More than 10% |
| (Sullman et al., 2015) | Buses | Static | Courses | / | Simulator, survey | Immediate, and 6-month after the training | 11.6% after the training, 16.9% fuel saving after 6 months |
| (Schall et al., 2016) | Logistics trucks | Static | Courses | Monetary and non-monetary incentives | On-road test | Immediatel, and 12-month after the training | Significant effect only when adding non-monetary incentives, while the effect fades afterwards |
| (Díaz-Ramirez et al., 2017) | Heavy- and medium-duty trucks | Static | Courses | / | On-road test | | A fuel reduction of 6.8% (in L/ton-100km) |

| (Barla et al., 2017) | Light-duty vehicles | Static | Courses | / | On-road test | 10-month after the training | Fuel saving of 4.6% on city roads and 2.9% on highway roads |
|---|---|---|---|---|---|---|---|
| (Jeffreys et al., 2018) | Light-duty vehicles | Static | Courses | / | On-road test | 12 weeks after the training | 4.6% fuel saving per 100km |
| (Wang and Boggio-Marzet, 2018) | Private vehicles | Static | Courses | / | On-road test | Immediate | On average 6.3% fuel savings ($CO_2$ reduction) |
| (Yao et al., 2019) | Light-duty vehicles | Static | Courses | / | Simulator | Immediate | 8.3% $CO_2$ reduction, 8.4% fuel saving |
| (Beloufa et al., 2019) | Light duty vehicles | Static | Courses, interactive guide | / | Simulator | Immediate | Up to 12.38% $CO_2$ reduction |
| (Goes et al., 2020) | Waste collection trucks | Static | Courses, coaches | / | On-road test | 3-month before and after the training | Up to US$18,507.55 per month of fuel cost saving, 7.1% reduction of $CO_2$-e emissions and local air pollutants |
| (Coloma et al., 2021) | Post vans | Static | Courses | / | On-road test, survey | 1-to-2 weeks after the training | Insignificant differences |

Note:

*this table only includes studies that measures energy saving and emission reduction

**unless noted, vehicles used in the experiments are fossil fuel powered

### 4.1.1   Performance on different vehicle types after static eco-driving training

Due to different vehicle powertrain and operation requirements, driving habits and associated behavioural change vary among drivers of different vehicle types after static eco-driving training. The change of $CO_2$ emissions and fuel efficiency was compared between gasoline and diesel vehicles in Wang and Boggio-Marzet (2018). Given the same segmentation and the same tier of emission standard, gasoline vehicles lead to higher fuel savings and more $CO_2$ reduction (-7.6%) than diesel vehicles (-4.7%). Barla et al. (2017) proved that manual transmission vehicles have a better performance than automatic transmission vehicles in fuel-saving on both city roads and highway roads after the eco-driving training. In addition, existing studies have also compared the difference in fuel consumption and emissions between gasoline vehicles and hybrid vehicles. For example, in Rutty et al. (2013), an eco-driving training program initiated in the city of Calgary, Canada, was analyzed, and they found that the daily $CO_2$ reduction of hybrid vehicles is smaller than that of gasoline vehicles by 0.5 kg. Comparing driving behaviour before/after the training, hybrid vehicles decrease average idling time by 10% per day, compared to a decrease of 4% from gasoline vehicles, while the count of harsh deceleration/acceleration of both vehicle types changes unnoticeably. Noted that in this study, the eco-driving course was explicitly developed for the municipality's fleet according to the local driving characteristics, instead of the general "Golden Rules" that are used in other studies.

### 4.1.2   Performance on different types of road after static eco-driving training

Existing studies have found that the energy-saving and emission reduction after eco-driving training sessions are more significant on congested road links, such as arterial and local streets, while on highway links where the speed is high and steady, the effect is less significant. Wang and Boggio-Marzet (2018) compared the driving behaviour on local roads, major arterials, and highway links after an eco-driving training in Madrid, Spain. The study observed that the behavioural change is more noticeable on major arterials with roundabouts and pedestrian crossings, especially the acceleration/deceleration rate (the absolute value reduced by 56.7% and 47.9%, respectively). The improvement of eco-driving behaviour enables the highest fuel saving on major arterial roads (8%) among all road types. Higher fuel savings on arterial links were also detected in Quebec, Canada: Barla et al. (2017) found that immediately after the training, the fuel consumption is reduced by 4.6% and 2.9% on city roads and highway roads, respectively. Andrieu and Pierre (2014) developed four driving indicators in response to the eco-driving guidelines and compare behavioural change under different speed limits. The time percentage of braking changes insignificantly when driving with a higher speed limit (90km/h). However, the fuel-saving variation due to different speed limits is not compared in this study.

### 4.1.3   Effects of static eco-driving trainings considering drivers' characteristics

Drivers' socio-demographic characteristics lead to heterogeneous effects on fuel consumption and emissions among the population, while current studies have not reached a common agreement on how these factors are associated with the effectiveness of eco-driving training. More specifically, Ho et al. (2015) analyzed the eco-driving training project in Singapore and found that male drivers have more notable behavioural changes with lower average speed after the training program, leading to higher fuel efficiency (15.98% versus 11.21% from female drivers) with more $CO_2$ reduction (-12.39% versus -9.02% from female drivers). Barla et al. (2017) applied a random coefficient model to assess the influence of an eco-driving pilot project in Quebec, Canada. Different from Ho et al. (2015), female drivers in Barla et al. (2017) have a higher probability to apply eco-driving techniques when driving on highway links. In addition, the model presents a large diversity in the change of fuel consumption among participants after the training session. However, this heterogeneity cannot be well-explained by the data. In Abuzo and Muromachi (2014), drivers from Tokyo, Japan and Manila, Philippines perform differently after receiving the same eco-driving course: Manila drivers showed a significant behavioural change on the stopping time and the number of stops in one trip, while the before/after difference of these two driving parameters was insignificant among Tokyo drivers. The study reveals the potential impact of nationality and associated cultural background, while the effect of drivers' socio-demographic factors on the result variation is not quantified. Díaz-Ramirez et al. (2017) applied a regression model to measure determinants on fuel consumption before and after an eco-driving training session towards truck drivers. Results show that driving experience greatly influences the effectiveness of the training, and with more experiences, fuel consumption is more difficult to reduce; however, other social norms such as age and education level do not have significant impact on fuel consumption either before or after the training.

Compared to private LDVs, drivers of buses and trucks are more experienced and professional, and they often operate vehicles on fixed driving routes and schedules with time constraints. In spite of the difficulty to change the well-trained behaviour of those professional drivers (Wu et al., 2018), the eco-driving training program is easier to implement because of regular tests and training sessions organized by companies or local authorities, and it is worth investigating the effect of the eco-driving training due to a remarkable share of emissions from these heavy-duty vehicles (Intergovernmental Panel on Climate Change, 2014). For example, Sullman et al. (2015) employed 29 bus drivers to attend an eco-driving course and evaluated their behavioural change using a driving simulator. A significant drop in fuel consumption (11.6%) was detected right after the training course. A follow-up survey was conducted again using the simulator after six months of the training, and the fuel-saving effect still existed. A significant reduction of $CO_2$ emissions and fuel consumption with the change of driving habits was also observed in Rutty et al. (2014), in which drivers of tourism buses were trained. Goes et al. (2020) trained drivers operating urban waste collection trucks, which are generally running in urban areas with much more stop-and-go operations compared to normal trucks. Their experiment revealed a significant reduction in $CO_2$ emissions (as high as 7.1%) after the training. Besides, environmental benefits were further revealed in this study by reducing the amount of air pollutant emissions. However, eco-driving training does not necessarily lead to improvements. Several studies have illustrated the difficulty to modify driving behaviour and habits, especially for these experienced employed drivers. For example, in Schall et al. (2016), without non-monetary rewards, professional drivers from a logistics company showed little behavioural change after receiving a theoretical eco-driving training session. Coloma et al. (2021) also revealed that courier delivery drivers are not likely to adopt eco-driving behaviour after the training due to the pressure of timely delivery.

### 4.1.4 Static eco-driving trainings with in-field instructions

Besides static eco-driving training sessions, some studies have tried to integrate a personalized coaching session with the training courses, with the hope that the in-field practice can efficiently improve the eco-driving behavioural. For example, Andrieu and Pierre (2014) compared the reduction of fuel consumption and emissions between two types of training sessions: theoretical teaching and in-field practice. The result shows that both formats can lead to the improvement of eco-driving behaviour, while the latter leads to lower average speed and higher obligation to the legal speed limit. Similar result was also depicted in Beloufa et al. (2019), in which the eco-driving rules were displayed on-board as the reminder for drivers. Yao et al. (2019) compared the result of traditional video training to personalized training, in which the instructions were provided based on individual characteristics. The result shows a customized training course can be more effective than a general training session. Wu et al. (2017) also concluded that personalized coaching improves fuel efficiency than a pure education session, while the behavioural change effect varies among different driving behaviours.

### 4.1.5 Sustaining eco-driving behaviour after static eco-trainings

Since eco-driving is not an automatized, natural, and "everyday" driving style (Pampel et al., 2018), whether drivers can maintain the eco-driving behaviour instead of returning to their "old habits" is essential to a sustainable environmental benefit of the training. First, interruptions during driving can lead to inconsistency of eco-driving behaviour. Pampel et al. (2018) conducted a driving experiment to test whether drivers can keep eco-driving behaviour after interruptions represented by a simplified Twenty Questions Test (Kafer and Hunter, 1997). The result shows that drivers cannot maintain eco-driving behaviour after the interruption, highlighting the difficulty of sustaining eco-driving behaviour after a static training session. Second, drivers may forget to practice eco-driving after the training for a long time.

In contrast to studies examining short-term effect right after the training (Rutty et al., 2013), several studies compared the short-term behavioural change with long-term driving habits (Allison and Stanton, 2019; Barla et al., 2017; Beusen et al., 2009), demonstrating a fading effect of the training along the time span. For example, Barla et al. (2017) assessed the fuel-saving effect of the eco-driving training immediately after the training session and after ten months of the session. The reduction of fuel consumption on arterial became 2.5% after ten months, compared to 4.6% immediately after the session.

Existing studies suggest regular alarm or periodic incentives may help maintain behavioural improvement, while the best method of the periodic alarm is yet to be determined. The effect of monetary incentives varies among experiments, and some researchers observe significant fuel savings when applying the reward for eco-driving. Lai (2015) tested monetary reward for eco-driving among bus drivers, which is provided if the actual fuel consumption is lower than the standard value. Through six-month experiment, the study found a long-term significant fuel saving (by 10%) if the reward is active. However, in Schall et al. (2016), a contradictory conclusion is drawn from a long-term experiment. This study tested the effect of eco-training over six months, during which they provided drivers with monetary incentives and non-monetary rewards based on their degree of eco-driving. The experiment found that the non-monetary reward leads to a short-term fuel saving, while it gradually weakens in a long run; however, neither long-term nor short-term improvement is observed in groups with monetary incentives. Similar conclusion can also be found in Díaz-Ramirez et al. (2017), which proves that providing driving feedback for a short-time experiment results in reduced fuel consumption, while long-term experiment shows an increase of fuel consumption among participants. These research poses the challenge to encourage drivers to continuously adopt eco-driving behaviour after the training, which is crucial to maintain environmental benefits from behavioural changes.

## 4.2 Dynamic guidance based on real-time driving operations

Static eco-driving training sessions deliver pre-determined eco-driving rules to drivers, and require drivers to apply them during real practices. The effectiveness of the learning process is questionable, and behavioural change towards eco-driving is hard to maintain for a long time (Alam and McNabola, 2014; Allison and Stanton, 2019). Instead, dynamic eco-driving guidance evaluates real-time driving performance based on driving operations, energy efficiency and emissions, and generates personalized driving suggestions to drivers via an in-vehicle or mobile driving assistance. The in-vehicle assistance device is composed of data sensors (such as OBD and GPS) and a module delivering the feedback to drivers. Dynamic eco-driving guidance can be developed in two methods. First, the system records and evaluates on-road driving behaviour, and sends a periodic personal driving report through in-vehicle devices or users' mobile devices. Second, the system generates instantaneous information, such as warnings or speed suggestions, and presents it to drivers through visualized, acoustic, or haptic notifications. In this way, real-time and specific feedback can be provided to drivers according to their own driving habits and on-road performance. Existing studies proposed different forms of the information module, varied by the type of feedback, and the method to deliver the information. Consequent energy saving and emission reduction depend on the acceptance and practice of drivers, which are related to not only the information type but also drivers' personality and psychological factors. In this subsection, different types of dynamic eco-driving guidance with examples of their application are introduced. The effect of dynamic eco-driving guidance in existing studies are presented in Table 2.

Table 2 Effects of dynamic eco-driving guidance on energy saving and emission reduction (ascending order of the published year)

| Study | Vehicle type* | Guidance type | Guidance design | Add-on options | Experiment type | Experiment duration | Energy saving/emission reduction effects |
|---|---|---|---|---|---|---|---|
| (Liimatainen, 2011) | Buses | Dynamic | Feedbacks | / | On-road test, survey | 1.5 years | 1.4%-4.6% fuel saving |
| (Ando and Nishihori, 2012) | / | Dynamic | Feedbacks | / | On-road test | Depends on the feedback frequency | Sporadic feedbacks leads to more $CO_2$ reduction than daily feedbacks |
| (Strömberg and Karlsson, 2013) | Buses | Dynamic | Visualized, coaches | / | On-road test, survey | 6 weeks | 6.8% fuel saving |
| (Vagg et al., 2013) | Light commercial vehicles | Dynamic | Visualized, auditory | / | On-road test | 2 weeks | On average 7.6% fuel saving |
| (Staubach et al., 2014) | Light duty vehicles | Dynamic | Visualized, haptic | / | Simulator | Immediate | On average 15.9% to 18.4% fuel saving |
| (Zhao et al., 2015) | Light duty vehicles | Dynamic | Feedbacks, visualized, auditory | / | Simulator | Immediate | 5.37% CO2 reduction, 5.45% fuel saving |
| (Dib et al., 2014) | Electric light duty vehicle | Dynamic | Visualized | / | On-road test | Immediate | 8.9% energy saving |
| (Caulfield et al., 2014) | Light duty vehicles | Dynamic | Feedbacks, visualized | / | On-road test | 10 months | On average 3% to 6% $CO_2$ reduction |
| (Rionda et al., 2014) | Bus | Static and dynamic | Courses and visualized advice | / | On-road test | 1 year | 7% fuel saving |

| (Orfila et al., 2015) | Light duty vehicles | Dynamic | Visualized | / | On-road test | Immediate | On average 30% fuel saving |
|---|---|---|---|---|---|---|---|
| (Rolim et al., 2016) | Light-duty vehicles | Dynamic | Feedbacks | / | On-road test | 3 months | 0.4% fuel saving, 9.3% $CO_2$ reduction |
| (Toledo and Shiftan, 2016) | Military vehicles | Dynamic | Feedbacks | / | On-road test | 50 weeks | 3-10% fuel saving |
| (Pozueco et al., 2017) | / | | Feedbacks | Peer-ranking | On-road test | 4 months | 31% fuel saving of the analyzed driver |
| (Wu et al., 2017a) | Taxi | Dynamic | Feedbacks | Peer-ranking | On-road test | 1 month | On average 4.5% fuel saving |
| (Wu et al., 2017c) | Taxi | Dynamic and static | Courses, coaching, feedback | Peer-ranking | Simulator, on-road test | 1 week | Up to 9.6% fuel saving |
| (Ayyildiz et al., 2017) | Trucks and light commercial vehicles | Dynamic and static | Courses, feedback, visualized | Peer-ranking | On-road test | 2 months | 5.5% fuel saving |
| (McIlroy et al., 2017b) | Light duty vehicle | Dynamic | Courses, visualized, auditory, haptic | / | Simulator, survey | Immediate | Up to around 22% fuel saving |
| (McIlroy et al., 2017a) | Light duty vehicles | Dynamic | Haptic | / | Simulator, survey | Immediate | 11% fuel saving |
| (McConky et al., 2018) | Commercial vehicle | Dynamic and static | Courses, feedbacks | Monetary rewards, peer-ranking | Simulator | Immediate | Peer competition has a more significant effect on CO2 reduction |

| (Ma et al., 2018) | Buses | Dynamic | Visualized, auditory | / | On-road test | 19-month in total | 6.25% fuel saving |
|---|---|---|---|---|---|---|---|
| (Zavalko, 2018) | Trucks | Dynamic | Visualized, coaches | / | On-road test | 3 months | 4% fuel saving |
| (Ko et al., 2018) | Light duty vehicles | Dynamic | Visualize | / | On-road test and simulator | Immediate | Up to 45% |
| (Vaezipour et al., 2019) | Light-duty vehicle | Dynamic | Visualized | Monetary reward, peer-ranking | Simulator | Immediate | 4.7% fuel saving |
| (Günther et al., 2020) | Electric light duty vehicles | Dynamic | Feedbacks | Monetary rewards, peer-ranking | On-road test, survey | 2-3 months | On average 1.02 to 2.99 kWh/100km energy saving |
| (Ng et al., 2021) | Light commercial vehicles | Dynamic | Auditory | / | On-road test | Immediate | 5%-6% fuel saving, up to 65% emission reduction (Nitrogen Oxide) |
| (Pinchasik et al., 2021) | Trucks | Static and dynamic | Courses, feedbacks | Non-monetary incentives | On-road test | One-year in total | 5.2% to 9% fuel saving |

Note:

*this table only includes studies that measures energy saving and emission reduction

**unless noted, vehicles used in the experiments are fossil fuel powered

### 4.2.1   Periodic reports and feedbacks

The periodic driving report usually includes the evaluation of driving behaviour and suggestions on the eco-driving improvement based on the real-driving records, and the report is pushed to the in-vehicle devices or users' mobile devices regularly. It can be regarded as a special type of eco-driving training, while the periodic report is more dynamic and customized to drivers' individual habits. Pinchasik et al. (2021) designed an eco-driving training with follow-up reports showing monthly eco-driving evaluations, which include a normalized score of average fuel consumption per 100km, engine and gear, speed, idling, deceleration, and cumulative distance. The result shows that participants reach fuel-saving of up to 9%, indicating the usefulness of regular driving evaluation, which may stimulate the awareness of potential eco-driving benefits. Another example of a monthly report in Pozueco et al. (2017) proposed a monthly driving feedback system that displays driving evaluation results on an in-vehicle dashboard. In this system, drivers' behaviour is classified into several efficiencies and safety levels, then both personal improvement (self-learning) and comparison among all participants (peer competition) are provided. The fuel-saving over 4 months was significant, which can reach 31%. Wu et al. (2017c) provided drivers with different types of eco-driving training and feedback methods according to their value and goal orientation, and the fuel saving effect can reach as high as 9.95%.

The frequency of the feedback is an essential factor that affects the learning process of drivers towards eco-driving skills. In Ando and Nishihori (2012), the experiment revealed that compared to daily feedback, sporadic feedbacks pushed monthly or weekly may have a better performance on alarming drivers to adopt eco-driving behaviour. A different result can be found in Wu et al. (2017a), which presents a negative linear relationship between fuel consumption and the frequency drivers checking the driving feedback. In contrast to Ando and Nishihori (2012), the result in Wu et al. (2017a) indicates the effectiveness of regular and frequent feedback. The content of feedback reports also matters. Rolim et al. (2016) examined the impact of delayed weekly or bi-weekly feedback on drivers' behavioural change. The short-term test shows that negative feedback (warnings of inappropriate behaviour) can be more effective in controlling aggressive driving behaviours compared to positive feedback (praises of appropriate behaviour). However, the long-term test shows increasing fuel consumption from both groups, and the reason is not clear in this study.

Similar to static training, the effect of additional incentives (monetary and non-monetary) on the eco-driving behaviour is tested with periodic dynamic feedbacks, and the result is also not consistent across different studies. For example, Günther et al. (2020) designed three schemes of dynamic eco-driving guidance, including an individual competition of eco-driving performance (intrinsic motivation), a financial incentive (extrinsic motivation), and a periodic driving feedback. Participants reported improved knowledge of eco-driving from all schemes, while the competition scheme leads to the highest fuel saving in field experiments. Financial reward, which is an extrinsic motivation, cannot further improve the result. This result is consistent with findings in other studies (for example, Schall and Mohnen (2015) and Schall et al. (2016)), while others found monetary incentives can be more effective than non-monetary counterparts (Bolderdijk et al., 2011; Lai, 2015). The variation of these results may be related to different testing participants and experiment designs, while the influencing factor as well as their effects have not been comprehensively investigated. To design a more effective and acceptable feedback system, studies focusing on the motivational factors and differences among various groups of drivers need to be conducted.

### 4.2.2 Gamification design in dynamic eco-driving guidance

The competition and reward design in Günther et al. (2020) refers to the concept of "gamification". It gradually attains attention from various disciplines due to its strength in experiential learning environments and higher satisfying level of users (Dos Santos et al., 2019; Günther et al., 2020). Gamification design provides intrinsic motivation and extrinsic motivation to users to improve their behaviour (Rodríguez et al., 2014), and the elements in existing applications are points (scores), progress feedback, and socialization (El Hafidy et al., 2021). Significant reduction of $CO_2$ emissions and energy consumption can be observed from drivers with gamified guidance or additional incentives (Günther et al., 2020; McConky et al., 2018; Vaezipour et al., 2019). However, the effect of gamification is debatable. Fitz-Walter et al. (2017) observe that a gamified design improved drivers' satisfaction towards the guiding system, while significant behavioural changes cannot be observed. Besides the doubt on its effectiveness, some researchers have expressed their worries about the negative impact of gamification: the behavioural change encouraged by monetary incentives cannot sustain upon removal (Yen et al., 2019), and the peer-competition through the ranking scheme may lead to overly competitive situation and cheating in practice (Liimatainen, 2011). In the current stage, gamification design used in driving guidance is mostly limited to traditional PBL (Points, Badge, and Leaderboard), while other elements and drivers' acceptance to these elements have seldom been investigated (Nacke and Deterding, 2017). A systematic design of gamified eco-driving guidance should be further explored, and the long-time effect of various gamified elements should be compared.

### 4.2.3 Dynamic eco-driving guidance with visualized suggestions

In a visualized system, eco-driving suggestions can be displayed on an in-vehicle digital device, and drivers are expected to adjust their driving behaviour according to the displayed information. The displayed information can be practical driving recommendations (such as the optimal speed and acceleration), warnings to improper behaviour (such as speeding, excessive braking, long-time idling), indicators showing the gap between current behaviour and the optimal one, and indicators of environmental performances (such as energy efficiency or emissions). For example, Ayyildiz et al.

(2017) added on-trip visualized driving feedbacks onto an eco-driving training session to aware drivers of real-time fuel consumption, emissions. The fuel saving can reach 5.5% after the training with the assistance of the visualized guidance.

However, existing studies also argue that the visualized system cannot lead to an effective change. In one interview of drivers who used visualized driving assistance, they stated that the eco-driving guidance is not practical and can be distractive (Fors et al., 2015). Caulfield et al. (2014) examined a real-time in-vehicle visualized driving warning system, and participants also receive periodic driving feedback. Although short-term benefits are observed, drivers tend to return back to their 'old habits'. The same results can also be depicted from Zavalko (2018), where truck drivers were guided dynamically with an in-vehicle visualized warning for excess energy usage, followed by a driving report. The study found a 13.6% fuel saving in the short-term, while the long-term effect decays to 4%. Strömberg and Karlsson (2013) used three colours to visualize the categorized gap between drivers' actual behaviour and suggested maneuvers. In addition to the insignificant benefits of using the visualized interface, the categorized driving performance is revealed to be difficult to keep, and the pressure of maintenance forces drivers to give up modifying their behaviour. The additional workload from reading visualized information raises safety concerns. In addition to the inherent shortage of visualized guidance, the display interface and the message design should be carefully designed. This indicates the need for a well-designed driving experiment to include different combinations of designs, which has not been sufficiently investigated.

### 4.2.4　Dynamic eco-driving guidance with combined sensory methods

In addition to the visualized presentation, existing research tried auditory notifications and haptic touch in dynamic eco-driving guidance with expectations of enhanced guidance effects. Vagg et al. (2013) designed a visual-audible warning device to provide a real-time eco-driving guide. Although the average fuel saving can reach 7.6%, the variation among individual drivers can range from 0.43% to 12.03%. Zhao et al. (2015) added voice prompts with the visual display during driving to warn non-eco-driving operations, the study observes an additional 2% fuel saving and CO2 reduction compared to the case with after-trip feedbacks only. Hammerschmidt and Hermann (2017) designs two types of auditory systems (metaphorical and continuous), and they are both more effective than the visualized display system in terms of fuel-saving.

In the experiment of McIlroy et al. (2017b), it was observed that auditory and haptic (by continuously vibrating the foot pedal) notifications are more effective than visualization on reducing drivers' aggressive behaviour such as speeding, harsh acceleration and braking. Another study from the same authors further tests the intensity and the length of time-to-event of the haptic stimulus. They found that the longest lead-time stimulus tends to be the most effective, while the users' perceived acceptance and the long-term effect of the haptic force still remain to be investigated (McIlroy et al., 2017a). A. H. Jamson et al. (2015) further distinguished haptic touch into two categories, forcing drivers to increase the pedal angle ('haptic force'), or adding resistance to prevent further pressing ('haptic stiffness'), and they found that the pedal angle is closer to the optimal level when implementing 'haptic force', compared to 'haptic stiffness'. Similar result is also depicted in Hibberd et al. (2015), that 'haptic force' leads to a better driving operation for deceleration rate compared to other dynamic guidance when the vehicle is supposed to decelerate; while for other operations such as cruising and acceleration, haptic touch did not outperforms other guidance types.

In terms of user acceptance, the auditory system is reported to be less disregarded than its visualized alternatives (Ahlstrom and Kircher, 2017), while its user experience is less satisfying (McIlroy et al., 2017b). Although eco-driving guidance through haptic touch shows less distraction, (Birrell et al., 2013; S. L. Jamson et al., 2015), safety concerns can be one of its shortages (Hibberd et al., 2015; S. L. Jamson et al., 2015). Through the comparison across studies using different sensory methods, we find that the results of behavioural change and speed control from such guidance system are not consistent. The influence of these sensory methods on users' acceptance and their environmental benefits need to be further discussed.

### 4.2.5   Optimized suggestions based on traffic condition identification

In the aforementioned dynamic eco-driving assistances, alarms and suggestions are pushed to drivers when their behaviour violates certain thresholds while surrounding traffic conditions of the target vehicle are not considered. Lacking considerations on traffic conditions lead to less traffic anticipation, and the guidance can be impractical to drivers, leading to a low acceptance level. In response to that, researchers integrate in-vehicle telecommunication systems to retrieve real-time traffic conditions, optimize instantaneous speed and acceleration, and generate eco-driving suggestions dynamically. Staubach et al. (2014) provided optimal speed suggestions based on the traffic signal to avoid idling and harsh acceleration. Although up to 18.4% fuel reduction can be reached and drivers reported a positive attitude, the study was implemented in a driving simulator without the consideration of surrounding vehicles. Ko et al. (2018) designed eco-driving speed guidance considering the downstream traffic signal, and the study developed an eco-signal system to adaptively change the traffic light. Both simulator and on-road test show a large fuel-saving potential (up to 41.9%), while the test was conducted at a single intersection with one testing vehicle, which is much simplified compared to the real-world.

Location information and associated road geometric design are adopted in the existing on-road test to identify traffic conditions by reading digital maps. The driving warnings/suggestions can be provided based on the optimal operation in specific traffic conditions (Dib et al., 2014; Heyes et al., 2015; Orfila et al., 2015). To identify the motion of other vehicles on-road, the lookahead camera has been integrated with the in-vehicle driving assistance to obtain real-time hazardous conditions and provide warnings to drivers (Ng et al., 2021). These applications lead to fuel reduction of 2.5% to 8.9%, with proof of the consistency between safe-driving and eco-driving. However, users' acceptance level and the long-term behavioral improvement still remain to be explored. Moreover, the optimal suggestions are generated theoretically with little consideration of drivers' individual preferences and driving habits. In response to that, Ma et al. (2018) developed an in-vehicle interface to display behavioural chain-based driving suggestions that are optimized by the shortest learning path. Fuel consumption can be reduced by 6.25% while minimizing the change of drivers' own driving style.

## 4.3 The additional workload from eco-driving and drivers' motivation

To investigate the cognitive load and acceptance towards different types of eco-driving assistance, existing studies conducted surveys to quantify the additional workload added from eco-driving instructions. Among these studies, the subjective mental workload is usually scored by the NASA Task Load Index (TLX) (Hart and Staveland, 1988a), and system acceptance can be measured by the System Acceptance Scale (SAS) of Van der Laan, which uses Usefulness and Satisfying as metrics (Hart and Staveland, 1988a). For example, Heyes et al. (2015) adopted the SAS to evaluate drivers' acceptance of the real-time driving advice provided by an in-vehicle system. The scale of Satisfying significantly improves after using the system, while the scale of Usefulness does not. The positive attitude towards eco-driving assistance also leads to less fuel consumption (-4.01%) compared to those who do not support the system. Hibberd et al. (2015) compared the workload among different types of dynamic sensory notifications, and the result from SAS and NASA-TLX indicates that the haptic system is less distractive with higher levels of usefulness and satisfaction than visual-auditory assistance.

In addition to survey-based experiments, real-time physiological measurement has been used in previous research to quantify the additional physiological workload from eco-driving. For instance, Ruscio et al. (2018) measured heart rate, blood volume pulse, and high frequency power of heart rate to quantitatively present the change of workload when providing in-vehicle eco-driving assistance to drivers who do not use it before. The measurement shows a higher cognitive load when drivers are asked for eco-driving, introducing potential driving risks. Ahlstrom and Kircher (2017) detected the glance behaviour with or without in-vehicle eco-driving guidance. The number and duration of glance vary when driving on different types of roads, while reduced mirror glances due to the in-vehicle guidance introduce more mental workload and higher risks during driving.

Limited studies investigated the inner motivation of drivers to implement driving behaviour. Existing research presents a large variation on the effect of eco-driving guidance among drivers from perspectives of both system acceptance and real-driving practice (Thijssen et al., 2014). The effect of socio-demographic factors, such as ages, genders, and education levels, varies across studies, and the result can be totally different (Delhomme et al., 2013; McIlroy and Stanton, 2017; Rolim et al., 2016). Lauper et al. (2015) describes the adoption of eco-driving as the consequence of two psychological processes, the formation of the behavioural intention, and the process of putting the intention into practice. They found that social norms only show minor effects on the intention of eco-driving; instead, drivers' attitude towards eco-driving and perceived behavioural control are the strongest predictor for eco-driving intention, and the action control is the strongest predictor for eco-driving practice. This study emphasizes that instead of socio-demographic factors, the inner motivation of drivers can be the most influential personal characteristic on eco-driving behaviour.

Environmental concerns, a positive attitude towards eco-driving and an open-mind to new technologies are possible inner motivation factors of drivers to practice eco-driving on-road (Caulfield et al., 2014; Stillwater et al., 2017). However, opposite results have also been found in existing studies. McIlroy and Stanton (2017) analyzed the relationship between the attitude towards eco-driving and the actual practice of eco-driving through an online survey. The result showed that the knowledge of eco-driving and awareness of environmental issues do not necessarily direct to eco-driving behaviour in real-world across genders, ages and levels of education. A similar result was also revealed in Scott and Lawson (2018), that drivers usually do not apply fuel-saving driving operations although they have related guidelines in mind., and the gap exists between the eco-driving knowledge and the practice. As was illustrated in Pampel et al. (2015), driving intervention is required to maintain the intention and put eco-driving into practice; otherwise, the drivers would not practice eco-driving behaviour in the real-world. The above-mentioned studies emphasize the challenge of encouraging eco-driving practices, which may need specific designs of eco-driving guidance systems, as well as possible involvement of intrinsic and extrinsic motivations with the consideration of psychological processes of human-being.

## 5 DISCUSSION AND FUTURE WORK

### 5.1 Review summary

In this paper, we categorize existing research about eco-driving guidance into two major types, static training and dynamic assistance, based on the utilization of real-time driving data to generate driving suggestions. Representative studies of both types are presented, and their effects on energy consumption and emissions are compared.

By comparing results of eco-driving experiments and surveys, we conclude that static eco-driving training cannot ensure a sustainable change of driving behaviour, while whether regular incentives help maintain the training effect or not is debatable across studies. As a 'semi-dynamic' guidance, the periodic driving report leads to energy saving and emission reduction in a longer time span than pure training, and the consequent behavioural change is affected by the frequency and the content of the report. Different sensory devices in dynamic eco-driving guidance are compared, and we find the haptic touch feedback is less distractive with higher users' acceptance than auditory and visualization; however, within the same sensory methods, eco-driving effects are inconsistent across studies. Comparing drivers' characteristics, inexperienced and unprofessional drivers are more likely to adopt driving suggestions and change their behaviour, although exceptions exist (see Huang et al. (2021)). Socio-demographic factors are associated with the adaptiveness of the eco-driving guidance while the influence is unclear and varied among studies. Most existing studies suggest that environmental concerns and a positive attitude towards eco-driving improve the acceptance level of eco-driving guidance, while drivers still need reminders to put their eco-driving knowledge into practice.

**5.2 Previous eco-driving review work**

Existing studies have reviewed current eco-driving research from multiple perspectives. From the aspect of decision making, Sivak and Schoettle (2012) systematically reviewed drivers' decisions ranging from vehicle choice, maintenance, route selection, and operational driving behaviour, and the study compared their effect on energy consumption; Alam and McNabola (2014) comprehensively investigated the claimed benefits of existing eco-driving policies and technologies and summarized the limitation of them.

From the aspect of technical methodologies, Zhou et al. (2016) summarized fuel consumption models that are suitable to be applied in eco-driving; Singh and Kathuria (2021b) reviewed driving behaviour profiling methods and their applications in eco-driving feedback. The Mintsis et al. (2020) reviewed existing research on eco-driving control optimization methods. Although some of the articles reviewed in Mintsis et al. (2020) implemented eco-driving advice to human drivers on the real road, this review emphasizes the application of the optimized control at signalized intersections.

Regarding review articles focusing on human-driver eco-driving guidance, Allison and Stanton (2019) summarized a wide range of eco-driving guidance systems, including eco-driving training and feedback. Their benefits and problems are qualitatively described. In addition, this work and similar review work such as Huang et al. (2018) and Sanguinetti et al. (2020) focused on the design of the guidance system, with highlights mostly on the fuel-saving effect of the design.

The research review of this paper distinguishes with previous review work from the following perspectives. First, this paper compared the effectiveness of static eco-driving in different scenarios, including varied drivers, vehicle types, road geometries, and sustaining timespans. Second, in addition to the effect of current eco-driving guidance, the effect of drivers' factors, including their acceptance, motivation to practice eco-driving, and their driving workload, are reviewed. Lastly, new types of system designs, such as the integration with V2X communication and the gamification elements, as well as their effect and drivers' attitude, are presented. Based on the novelties mentioned above, this paper poses current gaps of eco-driving guidance studies shown in the next section.

**5.3 Challenges**

Given a large amount of research on developing eco-driving guidance, we find the gap still exists between the effect of eco-driving guidance and variables that may affect the result. Eco-driving is not a natural driving style and has not become a general target in current driving training courses. Therefore, even if drivers may have eco-driving knowledge in mind (Pampel et al., 2015) or have received text messages about their eco-driving performance (Pampel et al., 2017), they would not implement eco-driving unless they are asked to. Current eco-driving guidance mostly plays the role of "reminding". The assistance system warns drivers of their inappropriate behaviour or provides basic suggestions, while how practical and instructive the information can be, is still under investigation. A standard agreement on the influencing factors of eco-driving guidance has not been reached. Many questions remain to be answered: for example, how socio-demographic characteristics influence the guidance result, which type of guidance is more suitable for a specific group of drivers, and how to generate adaptive driving suggestions according to instantaneous traffic conditions and personal driving habits.

Moreover, the long-term effect of eco-driving guidance remains debatable: most studies presented a fading effect of either static or dynamic eco-driving guidance, leading to an inconsistent energy-saving improvement (for example, Allison and Stanton (2019); Barla et al. (2017), Beusen et al. (2009), Rolim et al. (2016) and Zavalko (2018)). However, the requirement of the time span and the sample size for long-term effect measurement is much higher than that of instantaneous or short-term effect, and this possibly is the reason for the research gap of the long-term.

Another critical knowledge gap is the content of the eco-driving guidance that can maximally lead to behavioural change in real-world practice. Drivers report that they prefer simple, clear and informative eco-driving suggestions (Dahlinger et al., 2018; Fors et al., 2015; Sureth et al., 2019; Thill et al., 2018), and previous research indicated that the anticipation to traffic conditions (for example, gradually press the braking pedal when approaching a red-light signal) is more straightforward than other suggestions such as keeping a steady speed (Delhomme et al., 2013). However, the cruise control module has been gradually integrated into new vehicle models, and therefore, keeping a stable speed may no longer be a big problem. Currently, limited articles investigate drivers' perceived acceptance towards different behavioural guidance, and contents of driving suggestions have seldom been compared in real-world driving experiments, especially with the background of semi-automation and telecommunication technologies.

Although recent studies have proposed customized learning paths based on drivers' current driving behaviour to improve the acceptance level (Ma et al., 2018), the variation of tested traffic conditions, characteristics of participating drivers, as well as the adaptive driving guidance algorithm, still need to be investigated. In addition, a comprehensive experiment on different types of optimization algorithms is urged to maximize the effectiveness of the eco-driving guidance. On top of that, we suggest a standardized experiment design method so that the result of different studies can be comparable.

Considering safety issues, a large amount of current research utilized driving simulators for experiments (for example, Fors et al., 2015; Hibberd et al., 2015;

McConky et al., 2018; Pampel et al., 2018; Wu et al., 2018, 2017b). However, due to short driving distance, testing duration, and limited traffic scenarios, the driving simulator may not record enough influencing factors that reflect both inner motivation and learning progress of eco-driving behaviour. In addition, since drivers may modify their aggressive driving in purpose when recognizing the existence of the testing device, results from a driving simulator can be much better than that of on-road driving in terms of energy-saving and emission reduction. On-road tests are still necessary to develop an adaptive and easy-to-follow eco-driving assistance system, while the safety issue led by additional workload from reading eco-driving guidance should be considered. In future studies, driving simulator experiments can be used as a prior experience, followed by in-field studies, to capture more complicated traffic conditions and corresponding drivers' reactions. Future eco-driving guidance should also balance environmental benefits and driving safety to avoid potential accidents due to additional workload.

Lastly, the popularity of gamification introduces new presence formats of eco-driving guidance. Gamified elements commonly involved in existing designs are points and leaderboards, in other words, scores or ranking of their eco-driving behaviour among all the participants. However, due to the complexity of traffic conditions and the diversity of vehicle specifications, energy consumption levels can be incredibly varied among drivers who have similar driving habits, and the ranking-based scoring algorithm may be unfair. This further leads to the adaptation of the scoring algorithm and the acceptance of the gamified eco-driving guidance. Previous research has tried supervised learning to differentiate traffic conditions based on the geographic information and temporal factors of road segments (Xu et al., 2021), while the adaptiveness and scalability of the classification algorithm need to be improved.

Moreover, drivers also need to be classified in the ranking-based scoring system. In future studies, unsupervised learning algorithms implemented in risky behaviour detection such as Castignani et al. (2017) can be introduced to classify drivers' behaviour and identify traffic conditions around the targeted vehicle. In this case, a fair driving score specific for a particular group of drivers on certain driving conditions can be generated, which may adaptively encourage the behavioural change towards eco-driving.